\documentclass[10pt,twocolumn,letterpaper]{article}

\usepackage[pagenumbers]{cvpr} %

\usepackage{hanging} 
\usepackage{subfiles} %
\usepackage{makecell} %
\usepackage{multirow} %
\usepackage{subcaption}
\usepackage{color,xcolor}
\usepackage{graphicx}
\usepackage{booktabs}
\usepackage{makecell}
\usepackage{colortbl}
\usepackage{pifont}%
\usepackage{algorithm} %
\usepackage{algorithmic} %
\usepackage[switch]{lineno}
\usepackage{amssymb} %

\usepackage[misc]{ifsym}

\definecolor{shapecolor}{rgb}{0.0,0.5,0.0}

\newcommand{\boldparagraph}[1]{\vspace{0.2cm}\noindent{\bf #1}}

\definecolor{arylideyellow}{rgb}{0.91, 0.84, 0.42}

\definecolor{cvprblue}{rgb}{0.21,0.49,0.74}
\usepackage[pagebackref,breaklinks,colorlinks,citecolor=cvprblue]{hyperref}

\title{Masked Latent Transformer with the Random Masking Ratio to Advance the Diagnosis of Dental Fluorosis}

\author{Yun Wu$^{1,2,*}$, Hao Xu$^{1,2,*~\textrm{\Letter}}$, Maohua Gu$^{1,2}$, Zhongchuan Jiang$^{1,2}$, Jun Xu$^3$, Youliang Tian$^{1,2}$\\
$^{1}$ State Key Laboratory of Public Big Data, Guizhou University\\
$^{2}$ College of Computer Science and Technology, Guizhou University\\
$^{3}$ School of Artificial lntelligence, Nanjing University of Information Science and Technology\\
Code \& Dataset: \href{https://github.com/uxhao-o/MLTrMR}{ \ttfamily uxhao-o/MLTrMR}
}
\begin{document}

\maketitle

\let\thefootnote\relax\footnotetext{$^{*}$ Yun Wu and Hao Xu are co-first authors.\\$^{~\textrm{\Letter}}$ Corresponding author: Hao Xu (\url{uxhao\_o@163.com}).}

\begin{abstract}
    Dental fluorosis is a chronic disease caused by long-term overconsumption of fluoride, which leads to changes in the appearance of tooth enamel. It is an important basis for early non-invasive diagnosis of endemic fluorosis. However, even dental professionals may not be able to accurately distinguish dental fluorosis and its severity based on tooth images. Currently, there is still a gap in research on applying deep learning to diagnosing dental fluorosis. Therefore, we construct the first open-source dental fluorosis image dataset (DFID), laying the foundation for deep learning research in this field. To advance the diagnosis of dental fluorosis, we propose a pioneering deep learning model called masked latent transformer with the random masking ratio (MLTrMR). MLTrMR introduces a mask latent modeling scheme based on Vision Transformer to enhance contextual learning of dental fluorosis lesion characteristics. Consisting of a latent embedder, encoder, and decoder, MLTrMR employs the latent embedder to extract latent tokens from the original image, whereas the encoder and decoder comprising the latent transformer (LT) block are used to process unmasked tokens and predict masked tokens, respectively. To mitigate the lack of inductive bias in Vision Transformer, which may result in performance degradation, the LT block introduces latent tokens to enhance the learning capacity of latent lesion features. Furthermore, we design an auxiliary loss function to constrain the parameter update direction of the model. MLTrMR achieves 80.19\% accuracy, 75.79\% F1, and 81.28\% quadratic weighted kappa on DFID, making it state-of-the-art (SOTA). 
\end{abstract}

\begin{figure}[!t]
	\centering
	\includegraphics[width=\columnwidth]{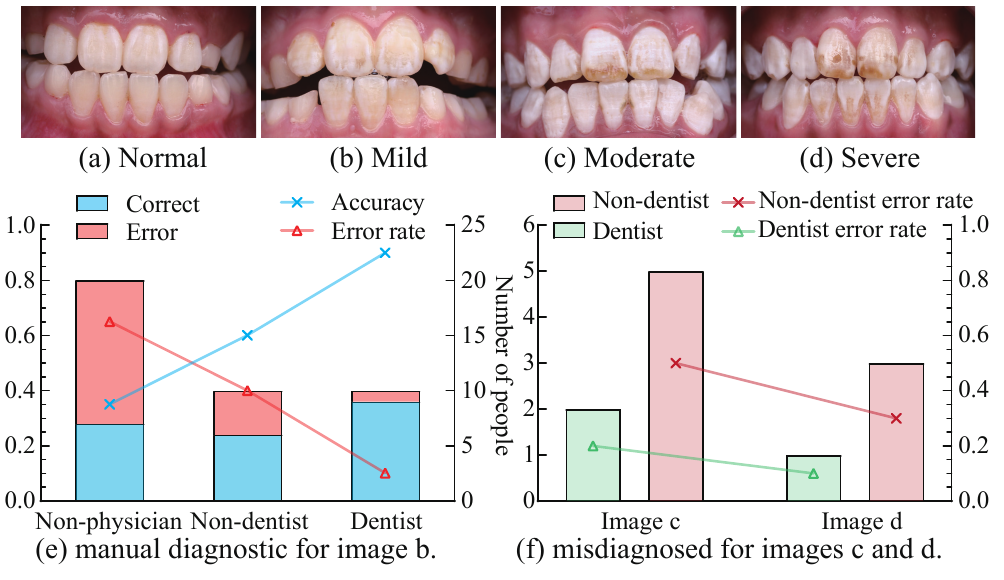}
	\caption{Comparison between normal dental and different severity degrees of dental fluorosis. (a) Normal. (b) Mild. (c) Moderate. (d) Severe. We invited 20 non-physicians(without medical or dental backgrounds), 10 non-dentists, and 10 dentists to manually assess images b, c, and d. Our statistics in (e) and (f) reveal the challenge faced in distinguishing between dental fluorosis and normal teeth. It's tough for non-physicians to distinguish them and non-dentists struggle to differentiate accurately between mild and moderate dental fluorosis. Notably, even dentists may not provide a completely accurate diagnosis.
 }
	\label{fig:xu1}
\end{figure}

\section{Introduction}
Dental fluorosis is a chronic disease caused by long-term overconsumption of fluoride, which leads to changes in the appearance of tooth enamel \cite{r1}. Also, it is an important basis for early non-invasive diagnosis of endemic fluorosis. As the severity of dental fluorosis changes, the surface of tooth enamel may show white spots, coloring, or even dental defects \cite{r2}. Excessive fluoride intake is mainly caused by excessive levels of fluoride in the air, soil, drinking water, and food in the natural environment \cite{r3}. Dental fluorosis is widespread worldwide, with China and India being the countries most affected by it \cite{r4,r5,r6}.

Dental fluorosis can have a range of effects, from aesthetic concerns to negative impacts on mental and physical well-being. Therefore, early detection of dental fluorosis is essential for effective prevention and treatment. Currently, the diagnosis of dental fluorosis depends solely on manual assessment by experienced dentists, whose expertise is crucial for precise diagnosis. Nevertheless, even dental professionals may not be able to accurately distinguish dental fluorosis and its severity based on tooth images (refer to Fig.~\ref{fig:xu1}). Moreover, there is a shortage of experienced professionals in regions with a high incidence of dental fluorosis, leading to a susceptibility to misdiagnosis and underdiagnosis when conducting large-scale dental fluorosis screening efforts. Underscoring the pressing requirement for an automated intelligent diagnostic approach.

As of April 2024, the literature search on the Web of Science reveals only four studies on the aided diagnosis of dental fluorosis. Reference \cite{r7} introduced an automated classification system for dental fluorosis utilizing fuzzy C-means clustering. Reference \cite{r8} presented a method to assess the severity of dental fluorosis by employing an enhanced unsupervised clustering algorithm based on K-means and fuzzy C-means algorithms within the HSV color model. Reference \cite{r9} applied an enhanced quantum-inspired fuzzy C-means clustering algorithm for dental fluorosis segmentation. Reference \cite{r10} suggested a particle optimization scoring K-nearest neighbor approach for predicting dental fluorosis. However, these studies rely on traditional machine learning techniques, assessing the severity of dental fluorosis solely through basic chromatic clustering models and the data used are proprietary.

In contrast to traditional methods, deep learning can automatically extract relevant features from vast amounts of data by learning patterns and features, providing greater robustness and generalization capabilities. The introduction of ResNet \cite{r11} has significantly advanced the use and development of deep learning models for various complex tasks. ResNet introduces residual learning, which solves problems such as gradient vanishing and explosion encountered in deep neural network training. DenseNet \cite{r12}, a densely connected convolutional network, establishes direct connections between each layer in the network, facilitating maximum information flow and solving gradient vanishing problems, leading to remarkable improvements in multiple target recognition tasks. SSAE \cite{r13} is a deep learning approach to nuclear detection in high-resolution breast cancer tissue pathology images, using pixel intensity to acquire sophisticated features and achieve efficient nuclear detection performance. Transformer \cite{r14} has made significant advances in natural language processing, and Dosovitskiy et al. have extended its application to computer vision by introducing Vision Transformer (ViT) \cite{r15}, providing new insights for medical image analysis. Swin Transformer \cite{r16} employs a self-attention mechanism to generate hierarchical representations of visual features, achieving outstanding results in image classification, object detection, and semantic segmentation tasks. MAE \cite{r17} is a scalable and effective masked autoencoder that focuses on reconstructing missing pixels by masking patches of input images with a fixed masking ratio and has strong generalization capabilities. The emergence of the U-Net \cite{r18} model, based on the architecture of Fully Convolutional Networks (FCNs) \cite{r19}, has significantly advanced the integration of deep learning in medical imaging. Despite the widespread application of deep learning in medical image analysis \cite{r20}, there is currently a lack of specific deep learning research dedicated to the grading diagnosis of dental fluorosis. A research gap persists in applying deep learning for diagnosing dental fluorosis. However, applying deep learning to diagnose dental fluorosis faces several challenges: \textbf{(i)} the absence of publicly available datasets containing dental fluorosis images for training deep learning models. \textbf{(ii)} the absence of deep learning models designed for automated diagnosis of dental fluorosis. \textbf{(iii)} potential variations in the quality of dental fluorosis images could impact the recognition performance of deep learning models.

To advance the development of deep learning in the assisted diagnosis of dental fluorosis, we construct the first open-source dental fluorosis image dataset (DFID) to lay the foundation for research in this field. Inspired by the above deep learning methods, we propose a pioneering deep learning model for automated intelligent diagnosis of dental fluorosis, called masked latent transformer with the random masking ratio (MLTrMR). MLTrMR introduces a mask latent modeling scheme based on Vision Transformer to enhance contextual learning of dental fluorosis lesion characteristics. Specifically, consisting of a latent embedder, encoder, and decoder, MLTrMR employs the latent embedder to extract latent tokens from the original image, whereas the encoder and decoder comprising the latent transformer (LT) block are used to process unmasked tokens and predict masked tokens, respectively. To mitigate the lack of inductive bias in Vision Transformer, which can lead to performance degradation, the LT block introduces latent tokens to enhance the learning capacity of latent lesion features. In particular, we observe that using a random masking ratio (0.3-0.8) for the masking operation achieves superior classification performance compared to a fixed masking ratio. Additionally, we design an auxiliary loss function to constrain the parameter update direction of MLTrMR. It minimizes the discrepancy between the feature map and the original image by reshaping the decoder output into a feature map matching the shape of the original image. It guides the model to update parameters in the optimal direction, significantly improving model performance. We create four model variants to investigate the impact of various hyperparameters on MLTrMR. On the dental fluorosis image dataset we constructed, MLTrMR achieves an accuracy of 80.19\%, an F1 score of 75.79\%, and a quadratic weighted kappa of 81.28\%, making it state-of-the-art (SOTA) for this excellent performance.

The main contributions are summarized as follows:

1) \hangpara{0.35in}{1}We construct the first open-source dental fluorosis images dataset (DFID), laying the foundation for research in deep learning to assist in the diagnosis of dental fluorosis. The establishment of DFID fills the data scarcity gap in this field and provides rich experimental resources for future research.

2) \hangpara{0.35in}{1}We take the lead in proposing a deep learning model for automated intelligent diagnosis of dental fluorosis, called masked latent transformer with the random masking ratio (MLTrMR). MLTrMR introduces a mask latent modeling scheme based on Vision Transformer, which enables automated intelligent diagnosis of dental fluorosis. Its introduction marked a pioneering step in integrating deep learning into diagnosing dental fluorosis, providing new insights and methods for further research.

3) \hangpara{0.35in}{1}We design an auxiliary loss function to constrain the parameter update direction of MLTrMR by reshaping the output of the decoder into a feature map matching the shape of the original image, thereby reducing the discrepancy between the feature map and the original image to guide the model towards optimal parameter updates, which significantly improves the model performance.

\section{Related Work}
\subsection{Medical Image Analysis Based on CNN}
Convolutional Neural Networks (CNNs) have been widely used in image processing due to their exceptional learning capabilities. The introduction of U-Net \cite{r18} has significantly influenced the application of deep learning in medical image analysis, laying the foundation for computer-aided medical diagnosis. U-Net++ \cite{r21} addresses the problem of lack of semantic information between the encoder and decoder by using a series of nested dense convolution blocks. Att U-Net \cite{r22} incorporates attention gates (AGs) to learn target structures autonomously, improving model sensitivity and accuracy by suppressing feature activation in irrelevant areas. FANet \cite{r23}, a feedback attention network, is specifically designed to improve medical image segmentation by effectively utilising the information gathered during each training epoch. Lin et al. \cite{r24} propose a novel lesion-based segmentation network that effectively exploits the distinction between internal and external features of the lesion area. Lie Ju et al. \cite{r25} propose a novel deep neural network method for learning retinal disease detection from a long-tail fundus database. Causal\_CXR \cite{r26} is a method that adopts a causal perspective to tackle the classification challenge of chest X-ray images.

\subsection{Medical Image Analysis Based on Vision Transformer}
Vision Transformer (ViT) \cite{r15} provides a new approach to medical image segmentation due to its outstanding ability to perceive global context. TransUNet \cite{r27} combines U-Net with ViT, using ViT as a powerful encoder to extract global context information and then using U-Net to restore local detail information, achieving significant performance improvements in several medical applications. SwinPA-Net \cite{r28} ingeniously integrates two carefully designed modules into the Swin Transformer architecture, resulting in the extraction of enhanced and robust features. ViT-AMCNet \cite{r29} is a network that employs adaptive model fusion and multi-objective optimization techniques for interpreting histopathological images related to the grading of laryngeal cancer tumors. 

The subsequent sections of this paper are organized as follows: Section III introduces the construction details of the fluorosis dental image dataset. Section IV provides a detailed description of the proposed MLTrMR model. Section V covers the model variations, data preprocessing processes, experimental parameters, evaluation metrics, and experimental settings. Section VI presents and analyzes the experimental results, and Section VII concludes this study.

\section{Dental Fluorosis Image Dataset}
The lack of high-quality, accurately annotated, open-source datasets is why deep learning cannot be applied to diagnosing dental fluorosis. Therefore, we aim to construct the first open-source, accurately annotated dental fluorosis image dataset (DFID) and lay the foundation for deep learning research in dental fluorosis diagnosis. 

\subsection{Dataset Collection and Annotation}
\begin{figure*}[!t]
	\centering
	\includegraphics[width=\textwidth]{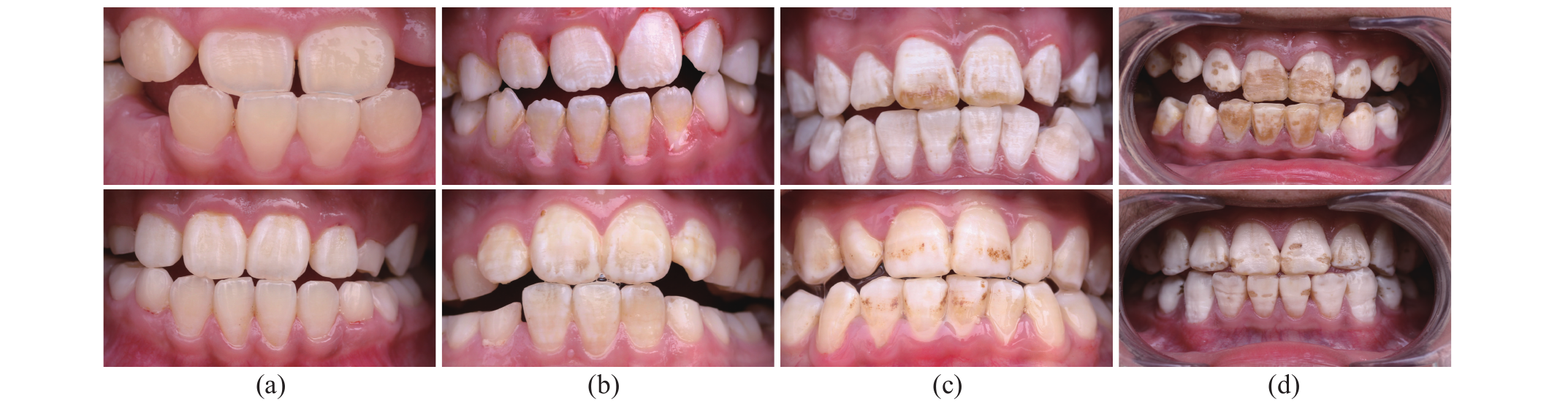}
	\caption{Example of dental fluorosis. (a), (b), (c), and (d) show normal dental, mild 
 dental fluorosis, moderate dental fluorosis, and severe dental fluorosis, respectively.}
	\label{fig:xu3}
\end{figure*}

We have collected data from patients with dental fluorosis in Guizhou Province, China. Currently, 131 participants have undergone comprehensive dental examinations, during which their teeth were photographed with an optical camera (Fig.~\ref{fig:xu3} shows part of the images). All participants have given their consent for us to use these data for scientific research. Professional dentists categorized the 131 dental images into four classifications based on the extent of fluorosis: normal, mild, moderate, and severe. The characteristics of each classification are shown in Fig.~\ref{fig:xu1}. Normal teeth have a translucent, milky-white appearance with smooth and shiny enamel. Mild dental fluorosis is characterized by white spots in the enamel and a loss of gloss on the tooth surface. Moderate dental fluorosis shows noticeable enamel surface wear and coloring, whereas severe dental fluorosis appears to have dental defects.

\begin{figure}[!t]
	\centering
	\includegraphics[width=\columnwidth]{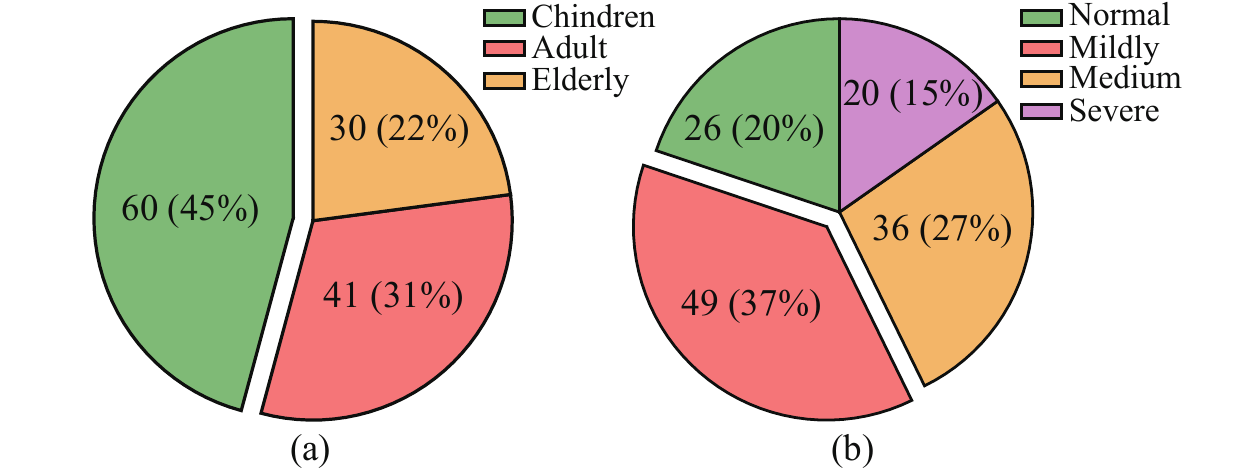}
	\caption{Data proportion statistics. (a) Distribution of age stages in participants. (b) Distribution of severity of dental fluorosis images.}
	\label{fig:xu2}
\end{figure}

\subsection{Dataset Statistics}
The DFID statistics are in Fig.~\ref{fig:xu2}, which includes 131 participants, including 60 children, 41 adults, and 30 elderly individuals. DFID consists of 131 dental images, each with a resolution of 560×448 pixels, categorized into four classifications: normal, mild, moderate, and severe. Specifically, there are 26 normal images, 49 mild images, 36 moderate images, and 20 severe images. The division of each classification into training and test sets follows a 7:3 ratio, as shown in Table~\ref{tab:1}.

\section{Methodology}
\subsection{Overall}
\begin{figure*}[!t]
	\centering
	\includegraphics[width=\textwidth]{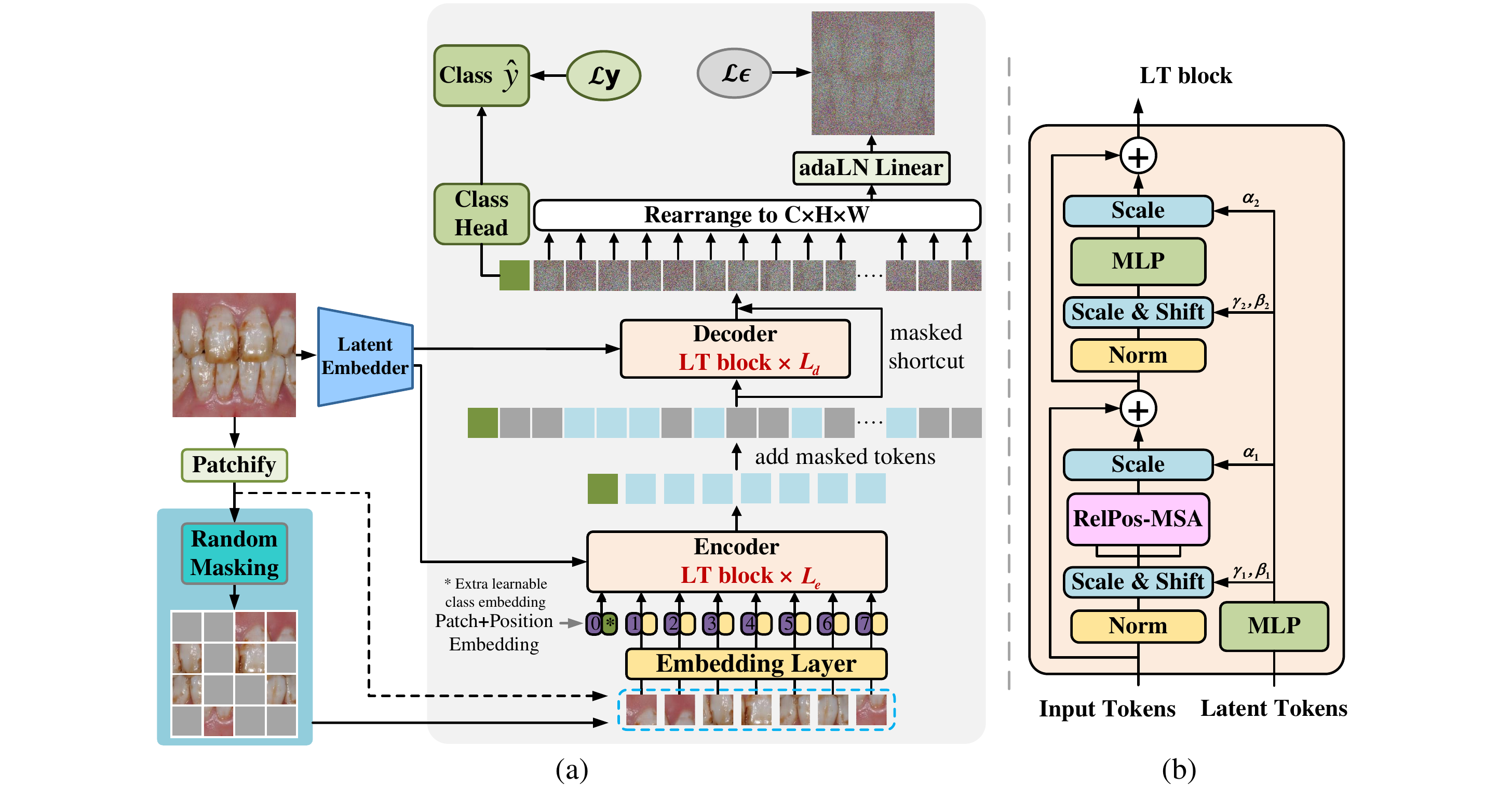}
	\caption{The overall structure of MLTrMR. (a) is the structure of MLTrMR. The embedding process before the encoder is similar to ViT. However, in our model, the masking operation is performed with a random ratio during training but is unnecessary during inference. The latent tokens produced by the latent embedder and the image tokens are fed into the latent transformer (LT) block. (b) is the LT block, which replaces the standard layer normalization layer in the Transformer block with adaptive layer normalization (adaLN), and regresses the scale and shift parameters $\gamma$ and $\beta$, as well as the dimension-wise scaling parameter $\alpha$ via the latent tokens. We design the relative position multi-head self-attention (RelPos-MSA) to replace the multi-head self-attention (MSA).}
	\label{fig:xu4}
\end{figure*}

To advance the diagnosis of dental fluorosis, we propose the masked latent transformer with the random masking ratio (MLTrMR). MLTrMR introduces a mask latent modeling scheme based on Vision Transformer to enhance contextual learning of dental fluorosis lesion characteristics. As shown in Fig.~\ref{fig:xu4}, MLTrMR consists of three components: latent embedder, encoder, and decoder. The latent embedder extracts latent tokens from the original image, whereas the encoder and decoder, consisting of the latent transformer (LT) block, process unmasked tokens and predict masked tokens, respectively. To mitigate the performance degradation due to the lack of inductive bias in ViT, we incorporate latent tokens in the LT block to enhance the learning capacity of latent lesion features. In addition, we integrate relative positional biases into multi-head self-attention to effectively capture the relationships between tokens. During training, a random masking ratio (0.3-0.8) is used to mask part of the patches from the input image, with the encoder processing only unmasked patches. The decoder takes the complete sequence generated by concatenating the encoder output with masked tokens as input and predicts the masked tokens from the unmasked tokens. To improve the model's adaptability to unknown information and noise, we introduce the masked shortcut before and after the decoder, ensuring that the decoder predicts masked tokens exclusively from unmasked tokens, thereby preserving the characteristics of the unmasked tokens in the final token sequence. Furthermore, the decoder's output is reshaped into a feature map matching the shape of the original image, and the mean squared error loss between this feature map and the original image is computed as an auxiliary loss function to constrain the parameter update direction of MLTrMR in training. During inference, no masking operation is performed (see the black dashed line in Fig.~\ref{fig:xu4}). Moreover, to output the classification results, we add a cls token to the header of all token sequences and classify them by a classification header after the decoder. The main components of MLTrMR are described in this section.

\begin{table}[!t]
    \caption{Training and Test Set Division Statistics in DFID}
    \centering
    \begin{tabular}{llll} 
        \hline
        Classification & Total & Training & Test  \\ 
        \hline
        Normal         & 26    & 18       & 8     \\
        Mild           & 49    & 35       & 14    \\
        Moderate       & 36    & 25       & 11    \\
        Severe         & 20    & 14       & 6     \\
        \hline
    \end{tabular}
    \label{tab:1}
\end{table}

\begin{figure}[!t]
	\centering
	\includegraphics[width=\columnwidth]{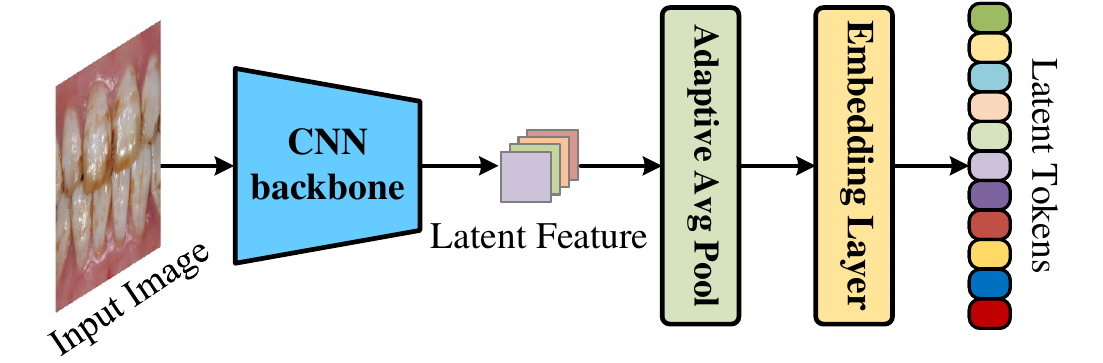}
	\caption{Structure of the latent embedder. The CNN backbone network extracts latent features from the original image, followed by dimensionality reduction through adaptive average pooling. The embedding layer then acquires latent tokens.}
	\label{fig:xu5}
\end{figure}

\subsection{Latent Embedder}
To mitigate the lack of inductive bias in Vision Transformer, which may result in performance degradation, we design a latent embedder, shown in Fig.~\ref{fig:xu5}, to extract latent tokens containing latent lesion features from the original image. Together, the latent tokens and image tokens serve as inputs to the LT block, designed to enhance the learning capacity for latent lesion features. The latent embedder consists of a CNN backbone, adaptive average pooling, and embedding layer. The CNN backbone extracts latent features from the original image, reduces dimensionality through adaptive average pooling, and produces latent tokens through the embedding layer. We use ResNet \cite{r14} or DenseNet \cite{r15} architectures without the fully connected layer (classification layer) as the CNN backbone.

\begin{figure}[!t]
	\centering
	\includegraphics[width=\columnwidth]{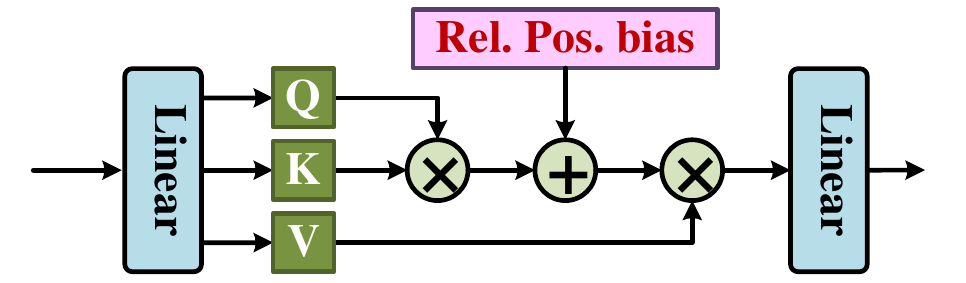}
	\caption{Structure of RelPos-MSA. Rel. Pos. bias denotes relative positional bias.}
	\label{fig:xu6}
\end{figure}

Assuming the input image is denoted as $\mathbf{x}\in\mathbb{R}^{H\times W\times C}$, where $(H, W)$ is its resolution and $C$ is the number of channels. The latent tokens $\mathbf{z}_{le}$ can be expressed as:
\begin{equation}\mathbf{z}_{le}=EL(AAP(CNN(\mathbf{x})))\in\mathbb{R}^{1\times D}\end{equation}
where $CNN$, $AAP$, and $EL$ denote the CNN backbone, adaptive average pooling, and embedding operation. $D$ is the dimension of the embedding layer.

\subsection{Masking Operation with the Random Masking Ratio} 
MAE \cite{r17} inspires our masking operation. In contrast to MAE's fixed masking ratio, we employ a random masking ratio within a predefined range, meaning each masking operation randomly selects a masking ratio from the specified range. Through experimentation, we have found that optimal performance is attainable when the random masking ratio ranges from 0.3 to 0.8. During the training, we implement random masking on the patch within the original image. Training the model on the unmasked patches enhances its capacity to learn from limited samples and improves its robustness. It is important to emphasize that the masking operation is exclusively performed during training.

Specifically, we reshap the image $\mathbf{x}\in\mathbb{R}^{H\times W\times C}$ into a sequence of flattened 2D patches $\mathbf{x}_{P}\in\mathbb{R}^{N\times(P^{2}\cdot C)}$, where $(H, W)$ is the resolution of the original image, $C$ is the number of channels, $(P, P)$ is the resolution of each patch, and $N=HW/P^2$ is the resulting number of patches. Subsequently, we perform the masking operation with a random masking ratio $\rho\in[0.3,0.8]$ to acquire unmasked patches $\mathbf{x}_{unp}\in\mathbb{R}^{N_1\times(P^2\cdot C)}$, where $N_1=\left\lfloor N\cdot(1-\rho)\right\rfloor$, and $\lfloor\cdot\rfloor$ denotes the floor function. And $\mathbf{x}_{unp}$ underwent an embedding operation to acquire unmasked tokens $\mathbf{z}_{unp}$. A binary mask vector $\mathbf{M}\in\mathbb{R}^N$ was created throughout this process, where 0 and 1 indicate masking and unmasking, respectively. Unmasked tokens $\mathbf{z}_{unp}\in\mathbb{R}^{N_1\times D}$ can be expressed as:
\begin{equation}
\mathbf{z}_{unp}=[\mathbf{x}_{unp}^{1}\mathbf{E};\mathbf{x}_{unp}^{2}\mathbf{E};...;\mathbf{x}_{unp}^{N_{1}}\mathbf{E}]
\end{equation}
where $\mathbf{E}\in\mathbb{R}^{(P^2\cdot C)\times D}$ is the patch embedding projection and $D$ is the dimensions of the embedding layer. $\begin{bmatrix}\cdot\end{bmatrix}$ is the concatenating operation. 

We generate a token for each input patch using embedding operations to streamline the processes above. Next, we randomly shuffle the token sequence and remove the last portion of the sequence using a random masking ratio. It guarantees that the unremoved tokens in the sequence correspond to patches randomly distributed throughout the original image. The unmasked tokens are subsequently concatenated to the encoder output and unshuffle (reverting the random shuffle operation) to align all tokens with their respective targets. (refer to Section IV.D for specifics on the decoder). This simple implementation introduces negligible overhead as the shufﬂing and unshuffling operations are fast. The above-simplified operation can be expressed as (3), equivalent to (2).
\begin{equation}
\mathbf{z}_{unp}=Masking_{\rho}(RS([\mathbf{x}_{p}^{1}\mathbf{E};\mathbf{x}_{p}^{2}\mathbf{E};...;\mathbf{x}_{p}^{N}\mathbf{E}]))
\end{equation}
where $Masking_\rho$ denotes the masking operation performed at a random masking ratio $\rho$. $RS$ denotes random shuffle. 

\subsection{Latent Transformer} 
As shown in Fig.~\ref{fig:xu4}, the LT block is the essential component of the encoder and decoder. We introduce latent tokens in the LT block to enhance the learning ability of latent lesion features, thereby avoiding the performance degradation caused by the lack of inductive bias in ViT. Specifically, inspired by DiTs \cite{r30}, we replace the standard layer normalization layer in the Transformer block with adaptive layer normalization (adaLN), regress the scale and shift parameters $\gamma$ and $\beta$, as well as the dimension-wise scaling parameter $\alpha$ via the latent tokens. As shown in Fig.~\ref{fig:xu6}, we enhance the multi-head self-attention mechanism by incorporating relative position biases, which we refer to as relative position multi-head self-attention (RelPos-MSA). The expression for RelPos-MSA is as follows:
\begin{equation}
\mathrm{\text{RelPos-MSA}(Q,K,V)=Softmax}\left(\frac{QK^{\mathrm{T}}}{\sqrt{d_{k}}}+B_{r}\right)\mathrm{V}
\end{equation}
where $Q$, $K$, and $V$ denote the query, key, and value. $d_{k}$ is the dimension of the key, and $B_r\in\mathbb{R}^{L\times L}$ is the relative positional bias that is selected by the relative positional difference $\Delta$ between the $i$-th position and other positions, $\operatorname{i.e.}B_r(i,\Delta)$. $L$ is the length of the sequence input into RelPos-MSA.

\subsubsection{Encoder}
The encoder comprises the $L_{e}$ layer LT block. During the training, the unmasked patches undergo the same embedding operation as ViT to acquire unmasked tokens $\mathbf{z}_{unp}$ (for simplicity, we perform (3)). Subsequently, both $\mathbf{z}_{unp}$ and latent tokens $\mathbf{z}_{le}$ are fed into the encoder. However, during the inference, the encoder handles the complete token sequence. Furthermore, akin to ViT, a cls token $\mathbf{x}_{cls}\in\mathbb{R}^{1\times D}$ is appended at the header of the token sequence. We also add positional embeddings for these tokens. The execution process of the encoder can be expressed as (5)-(8).
\begin{equation}\mathbf{z}_{0}=[\mathbf{x}_{cls};\mathbf{z}_{unp}^{1};\mathbf{z}_{unp}^{2};...;\mathbf{z}_{unp}^{N_{1}}]+\mathbf{E}_{pos}\end{equation}
\begin{equation}\gamma_{1},\beta_{1},\alpha_{1},\gamma_{2},\beta_{2},\alpha_{2}=\mathrm{MLP}(\mathbf{z}_{le})\end{equation}
\begin{equation}\mathbf{z}_{l}^{\prime}=\alpha_{1}\cdot\text{RelPos-MSA}((1+\gamma_{1})\cdot\text{LN}(\mathbf{z}_{l-1})+\beta_{1})+\mathbf{z}_{l-1}\end{equation}
\begin{equation}\mathbf{z}_{l}=\alpha_{2}\cdot\mathrm{MLP}((1+\gamma_{2})\cdot\mathrm{LN}(\mathbf{z}_{l}^{\prime})+\beta_{2})+\mathbf{z}_{l}^{\prime}\end{equation}
where $\mathbf{E}_{pos}\in\mathbb{R}^{(N_{1}+1)\times D}$ is the positional embeddings. $l=1...L_e$ denotes layer $l$ in the encoder. $\gamma$, $\beta$, and $\alpha$ denote the scale, shift, and the dimension-wise scaling parameter, respectively.

\subsubsection{Decoder}
The decoder comprises the $L_{d}$ layer LT block. During the training, randomly initialized masked tokens $\mathbf{z}_m\in\mathbb{R}^{(N-N_1)\times D}$ are appended to the output of the encoder $\mathbf{z}_{L_e}$, followed by an unshuffle operation to align all tokens with their target, resulting in $\mathbf{z}_{0}\in\mathbb{R}^{(N+1)\times D}$. Both $\mathbf{z}_{0}$ and $\mathbf{z}_{le}$ are fed into the decoder to predict $\mathbf{z}_{m}$. The execution process of the decoder can be expressed as (9)-(12).
\begin{equation}\mathbf{z}_0=UnRS([\mathbf{z}_{L_e};\mathbf{z}_m])\end{equation}
\begin{equation}\gamma_{1},\beta_{1},\alpha_{1},\gamma_{2},\beta_{2},\alpha_{2}=\mathrm{MLP}(\mathbf{z}_{le})\end{equation}
\begin{equation}\mathbf{z}_l^{\prime}=\alpha_1\cdot\text{RelPos-MSA}((1+\gamma_1)\cdot\text{LN}(\mathbf{z}_{l-1})+\beta_1)+\mathbf{z}_{l-1}\end{equation}
\begin{equation}\mathbf{z}_{l}=\alpha_{2}\cdot\mathrm{MLP}((1+\gamma_{2})\cdot\mathrm{LN}(\mathbf{z}_{l}^{\prime})+\beta_{2})+\mathbf{z}_{l}^{\prime}\end{equation}
where $UnRS$ denotes unshuffle. $l=1...L_d$ denotes layer $l$ in the decoder. $\gamma$, $\beta$, and $\alpha$ denote the scale, shift, and the dimension-wise scaling parameter, respectively.

\subsubsection{Masked shortcut}
To improve the model's adaptability to unknown information and noise, we introduce the masked shortcut before and after the decoder, ensuring that the decoder predicts masked tokens exclusively from unmasked tokens, thereby preserving the characteristics of the unmasked tokens in the final token sequence. The masked shortcut can be expressed as (13)-(15).
\begin{equation}\mathbf{z}_{in}=[\mathbf{z}_{0}^{1};\mathbf{z}_{0}^{2};...;\mathbf{z}_{0}^{N}]\end{equation}
\begin{equation}\mathbf{z}_{out}=[\mathbf{z}_{L_{d}}^{1};\mathbf{z}_{L_{d}}^{2};...;\mathbf{z}_{L_{d}}^{N}]\end{equation}
\begin{equation}\mathbf{z}=[\mathbf{z}_{L_{d}}^{0};\mathbf{M}\cdot\mathbf{z}_{\mathrm{in}}+(1-\mathbf{M})\cdot\mathbf{z}_{out}]\end{equation}
where $\mathbf{z}_{0}$ is the same as $\mathbf{z}_{0}$ in (9). $\mathbf{M}$ is the binary mask vector defined in Section IV.C. $\mathbf{z}_{in}$ and $\mathbf{z}_{out}$ are the tokens remaining after removing the cls tokens $\mathbf{x}_{cls}$ from the header at input $\mathbf{z}_0$ and output $\mathbf{z}_{L_d}$ of the decoder, respectively.

The final output of the MLTrMR has a prediction classification of $y^{\prime}$.
\begin{equation}y^{\prime}=\text{CH}(\mathbf{z}^0)\end{equation}
where $\mathbf{z}^0$ denotes the cls token in the header of the decoder output sequence. $\text{CH}$ is a class head.

\subsection{Auxiliary Loss Function}
MAE \cite{r17} is a method used for image reconstruction, falling within the realm of image generation. MAE reconstructs the original image by learning latent features and predicting masked tokens. This learning process involves minimizing the difference between the generated and actual tokens. The model continuously adjusts the generated tokens to match the target tokens, directing the parameter updates accordingly. We assume that aligning the image features in the output token sequence with the original image features can guide the model optimization and enhance performance. Experimental results confirm this assumption. Specifically, we reshape the decoder output into a feature map $\mathbf{x}^{\prime}\in\mathbb{R}^{H\times W\times C}$, matching the shape of the original image $\mathbf{x}\in\mathbb{R}^{H\times W\times C}$, and minimizing the mean-square error $\epsilon$ between them can facilitate optimal parameter updates during training. We then add $\epsilon$ as the auxiliary loss function to the cross-entropy loss (CELoss) function, and CELoss calculates the discrepancy between the predicted and the true class $y^{\prime}$ and $y$. The final loss function is:
\begin{equation}L=\mathrm{CELoss}(y^{\prime},y)+\frac{1}{H\times W\times C}\big\Vert\mathbf{x^{\prime}}-\mathbf{x}\big\Vert_{2}^{2}\end{equation}

\section{Experiment Setup}
\subsection{Model Variants}
We referred to the configurations of ViT \cite{r18} and Swin Transformer \cite{r19} to establish the base model MLTrMR-B, the small model MLTrMR-S, the large model MLTrMR-L, and the huge model MLTrMR-H. The detailed parameters of these models are presented in Table~\ref{tab:2}, while the model sizes and theoretical computational complexities of these variants are outlined in Table~\ref{tab:4} and Table~\ref{tab:5}. It is essential to highlight that the objective behind creating these variants is to delve deeper into the influence of various hyperparameters on the performance of the MLTrMR model.

\subsection{Data Preprocessing}
\begin{figure}[!t]
	\centering
	\includegraphics[width=\columnwidth]{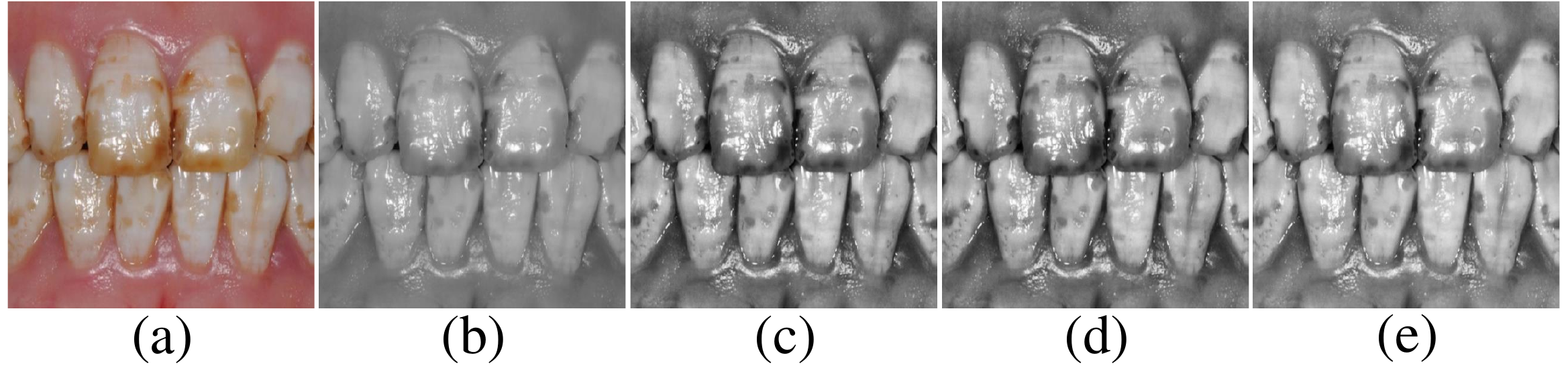}
	\caption{The process of preprocessing. (a) Original image. (b) Grayscaling. (c) Normalization. (d) CLAHE (e) Gamma Correction $\gamma=1.2$.}
	\label{fig:xu7}
\end{figure}

We use fuzziness, flipping, shifting, scaling, and 90° rotation to augment the training set, resulting in a training set that is 64 times larger than the original training set. In addition, the resolution of all images is reshaped to 512×512. When capturing dental images of fluorosis, achieving consistent color across all images poses a challenge due to variations in lighting, shooting angles, and other factors. Consequently, specific images may exhibit inadequate contrast in the lesion regions. Data preprocessing is essential before model training to mitigate the influence of these variables on model training. As shown in Fig.~\ref{fig:xu7}, this preprocessing encompasses grayscaling, normalization, contrast-limited adaptive histogram equalization (CLAHE), and gamma correction. 

\subsection{Implementation Details}
The MLTrMR was implemented using PyTorch 2.0 and experimented on a single NVIDIA RTX A6000 (48GB) GPU. We employed the Ranger optimizer. The learning rate was set to 1e-4, and we utilized a cosine annealing strategy with a weight decay of 1e-5 to dynamically adjust the learning rate. The batch size was set to 16. Maximum number of training is 50 epochs.

\begin{table}[!t]
    \caption{Details of MLTrMR Model Variants}
    \centering
    \includegraphics[width=\columnwidth]{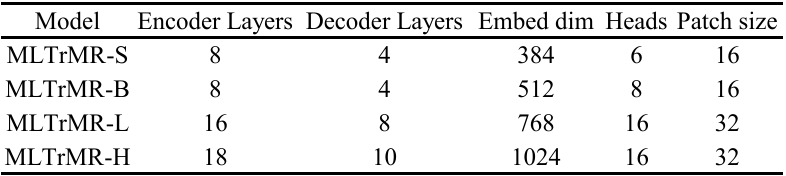}
    \label{tab:2}
\end{table}

\subsection{Evaluation Metrics}
We evaluated the proposed MLTrMR model during the experiment using three evaluation metrics: accuracy (Acc), F1 score, and quadratic weighted kappa (qw Kappa). The quadratic weighted kappa is as follows:
\begin{equation}\begin{aligned}
&\mathrm{qw}Kappa=\mathrm{l}-\frac{\sum_{i,j}W_{i,j}\cdot O_{i,j}}{\sum_{i,j}W_{i,j}\cdot E_{i,j}} \\
&W_{i,j} =\frac{\left(i-j\right)^2}{\left(N-1\right)^2}  \\
&E_{i,j} =\frac{\sum_{k}O_{i,k}\cdot\Sigma_{k}O_{k,j}}{\Sigma_{i,j}O_{i,j}} 
\end{aligned}\end{equation}
where $W_{i,j}$ is square weighted. $O_{i,j}$ is an element in the confusion matrix. $E_{i,j}$ denotes the element of the expectation matrix, and $N$ is the number of classifications.

\section{Experiment Results and Discussion}
\subsection{Comparisons With the Other Methods}
We compared the performance of MLTrMR with the classical models ResNet \cite{r14}, DenseNet \cite{r15}, ViT \cite{r18}, and Swin Transformer \cite{r19}. It is worth noting that we did not compare with all deep learning models mentioned in the introduction and related work sections for several reasons: \textbf{(i)} some models used for medical images are for segmentation tasks, while our research focuses on classification; \textbf{(ii)} these models are designed for specific tasks and are not suitable for dental fluorosis diagnosis; \textbf{(iii)} they have not made their code publicly available, so we cannot fully reproduce these models. 

\begin{figure*}[!t]
	\centering
	\includegraphics[width=\textwidth]{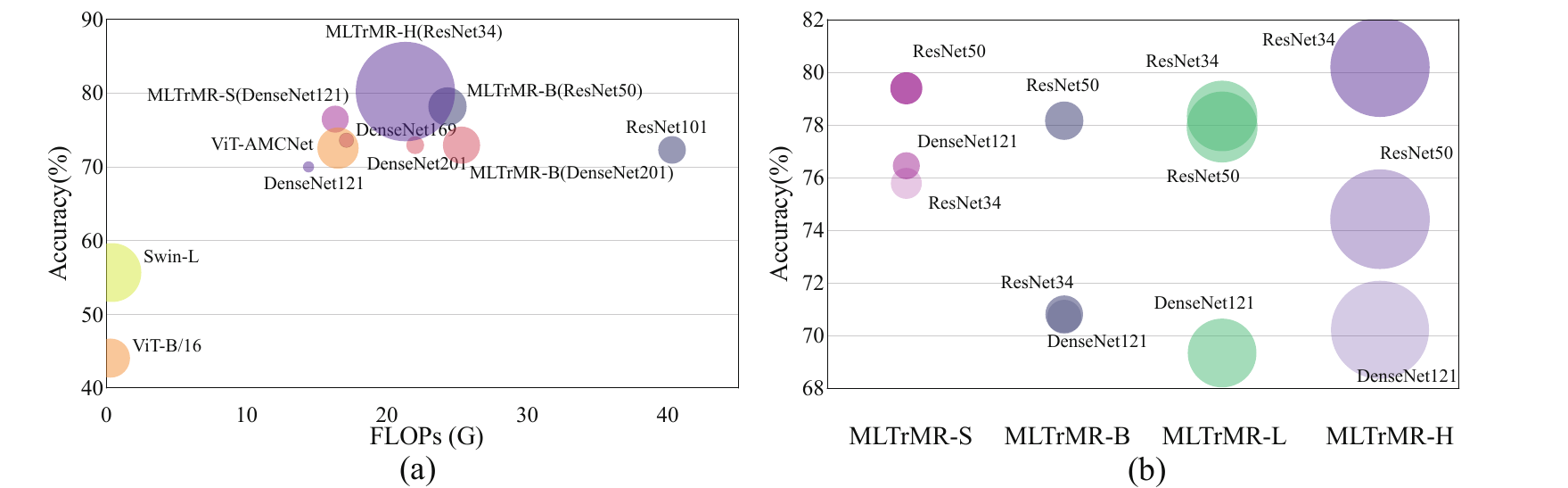}
	\caption{Comparison of MLTrMR variants with other models. The bubble area indicates the number of parameters of the model. (a) By comparing MLTrMR with other methods, MLTrMR achieves the best accuracy. (b) Comparison of MLTrMR variants of latent embedder consisting of different backbones with up to 80\% accuracy surpassed.} 
	\label{fig:xu8}
\end{figure*}

Fig.~\ref{fig:xu8}(a) and Table~\ref{tab:3} present the experimental results compared to other methods. When ResNet serves as the backbone for the latent embedder, MLTrMR demonstrates significant enhancements in Acc, F1, and qw Kappa compared to ResNet, with improvements ranging from 5.88\% to 7.91\%, 1.63\% to 7\%, and 0.09\% to 2.44\%, respectively. Similarly, when DenseNet is utilized as the backbone network for potential embeddings, MLTrMR also exhibits notable improvements in Acc, F1, and qw Kappa compared to DenseNet, with enhancements ranging from 1.08\% to 2.06\%, 1.97\% to 4.53\%, and 0.45\% to 6.19\%, respectively. Moreover, the MLTrMR model shows strong performance with parameters and computational complexity moderate to add. Our proposed MLTrMR significantly outperforms ViT and Swin Transformer. The MLTrMR model is robust, as incorporating random masking operations during training enables the model to discern the disparities between fluorosis and dental lesion images with minimal features. In contrast, ViT and Swin Transformer can solely learn from full features, leading to subpar classification performance on the DFID and a lack of robustness.

\begin{table}[!t]
    \caption{Comparison of MLTrMR with Other Models}
    \centering
    \includegraphics[width=\columnwidth]{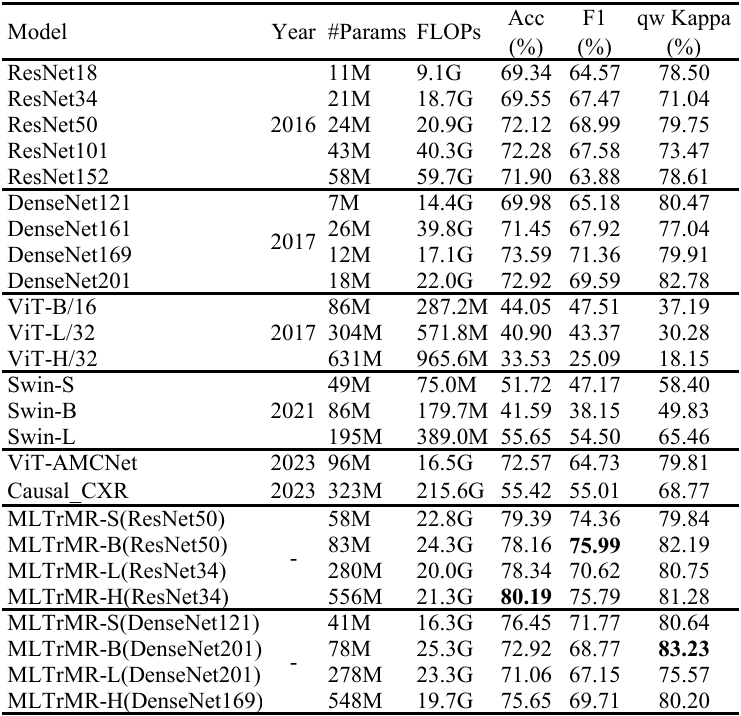}
    \label{tab:3}
\end{table}

\begin{table}[!t]
    \caption{Comparison of MLTrMR Variants (ResNet as the Backbone)}
    \centering
    \includegraphics[width=\columnwidth]{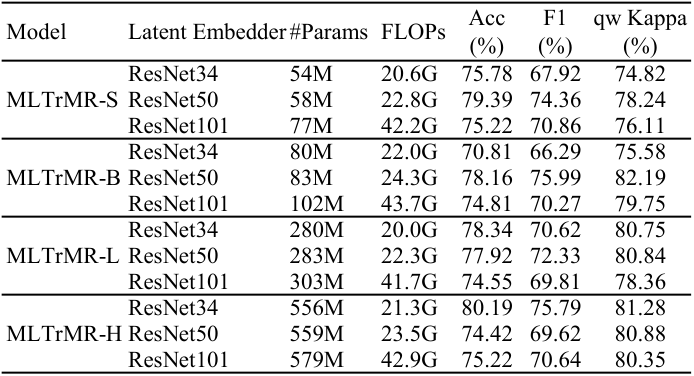}
    \label{tab:4}
\end{table}

\subsection{Comparison of MLTrMR Variants}
To investigate the impact of hyperparameters and different backbones in the latent embedder on model performance, we combined MLTrMR variants with the latent embedder consisting of different backbones, exploring the optimal combination. Specifically, We considered ResNet (ResNet34, ResNet50, ResNet101) and DenseNet (DenseNet121, DenseNet169, DenseNet201) as the backbones in the latent embedder. The experimental results are shown in Fig.~\ref{fig:xu8}(b), Table~\ref{tab:4}, and Table~\ref{tab:5}. When ResNet and DenseNet are used as the latent embedder's backbone, different MLTrMR variants can achieve excellent performance. The experimental results indicate our MLTrMR is insensitive to hyperparameters.

\subsection{Ablation Studies}
We conducted ablation studies from multiple perspectives to validate the effectiveness of each component of MLTrMR.

\boldparagraph{Impact of the Pre-trained Backbone and Freezing Parameters in the Latent Embedder.} To examine the impact of utilizing pre-trained backbones and freezing parameters in the latent embedder on the MLTrMR model's performance, we conducted experiments on two variants: MLTrMR-B with ResNet50 and ResNet34 as the backbone and MLTrMR-S with DenseNet121 and DenseNet169 as the backbone. The results of the experiments are detailed in Table~\ref{tab:6}. Our findings indicate that employing pre-trained backbones led to enhanced model performance. Specifically, there were notable improvements in ACC, F1, and qw kappa metrics ranging from 0.24\% to 9.79\%, 1.13\% to 8.41\%, and 2.35\% to 11.14\%, respectively. The backbones were pre-trained on DFID, enabling them to capture effective latent features across various lesion levels. This equips the MLTrMR model with valuable latent tokens for learning lesion features, enhancing model performance. Furthermore, our observations suggest that freezing the parameters of the pre-trained backbone may not yield optimal results. This limitation could stem from the static nature of the learned latent features when the backbone parameters are frozen, hindering adaptive adjustments according to training requirements and resulting in performance degradation.

\begin{table}[!t]
    \caption{Comparison of MLTrMR Variants (DenseNet as the Backbone)}
    \centering
    \includegraphics[width=\columnwidth]{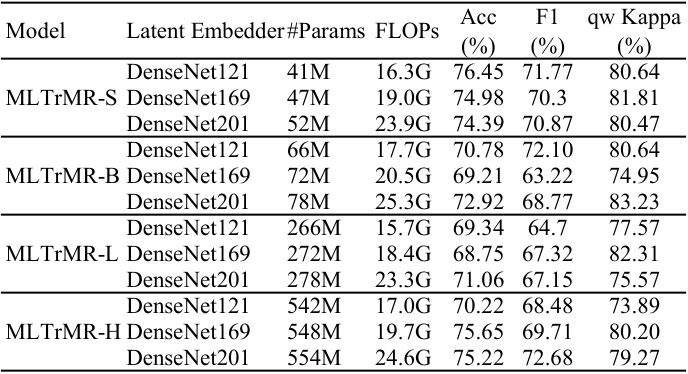}
    \label{tab:5}
\end{table}

\begin{table*}[!t]
    \caption{Ablation Studies on the MLTrMR-B(ResNet50)}
    \centering
    \includegraphics[width=\textwidth]{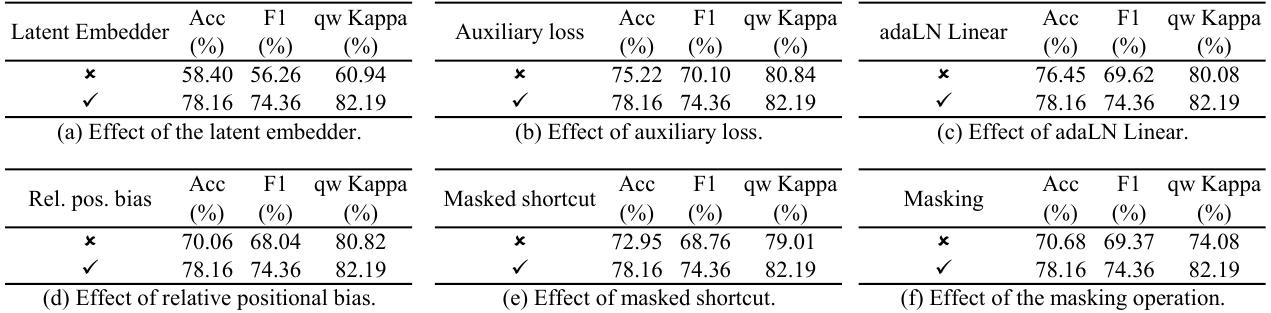}
    \label{tab:8}
\end{table*}
\begin{table}[!t]
    \caption{Effect of the Pre-trained Backbone and Freezing Parameters on MLTrMR}
    \centering
    \includegraphics[width=\columnwidth]{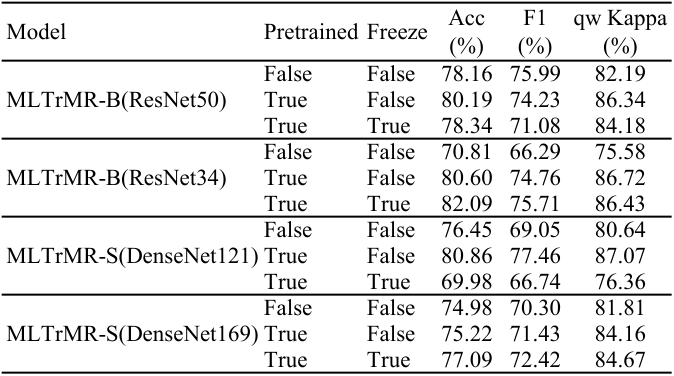}
    \label{tab:6}
\end{table}

\begin{figure}[!t]
	\centering
	\includegraphics[width=\columnwidth]{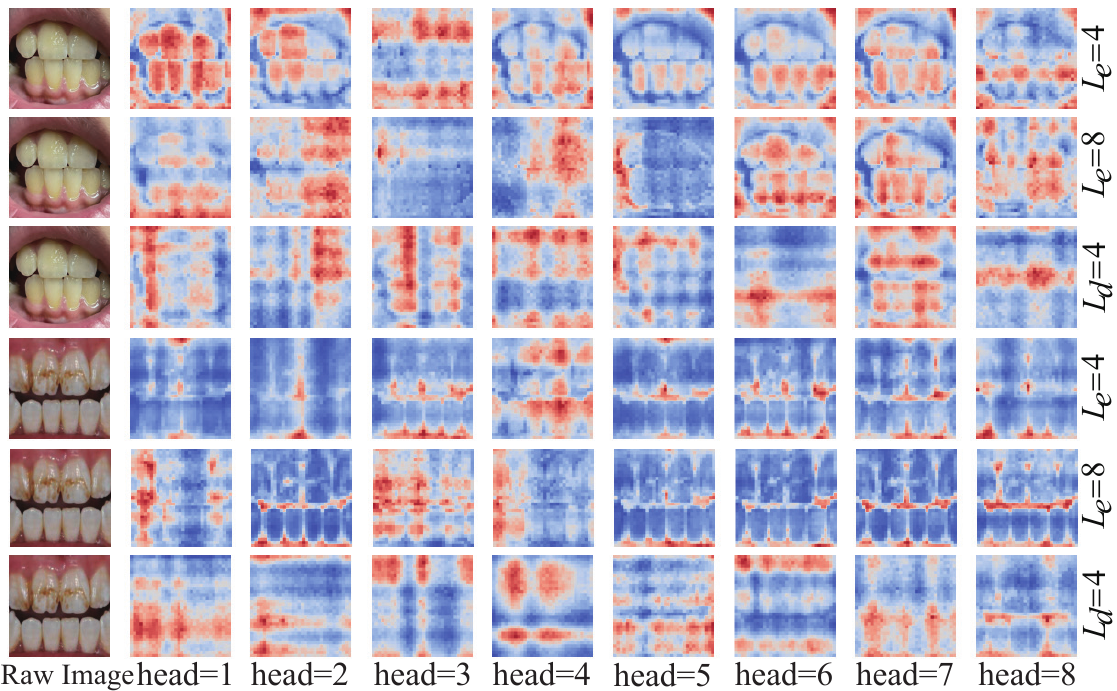}
	\caption{Visualization of attention in RelPos-MSA on MLTrMR-B(ResNet50). The figure displays the original image in the first column, followed by the attention of each of the eight heads. The three consecutive lines from top to bottom represent the 4th and last layer of the encoder and the last layer of the decoder.} 
	\label{fig:xu9}
\end{figure} 

\begin{table}[!t]
    \caption{Comparison of Various Masking Ratios and the Random Masking Ratio}
    \centering
    \includegraphics[width=\columnwidth]{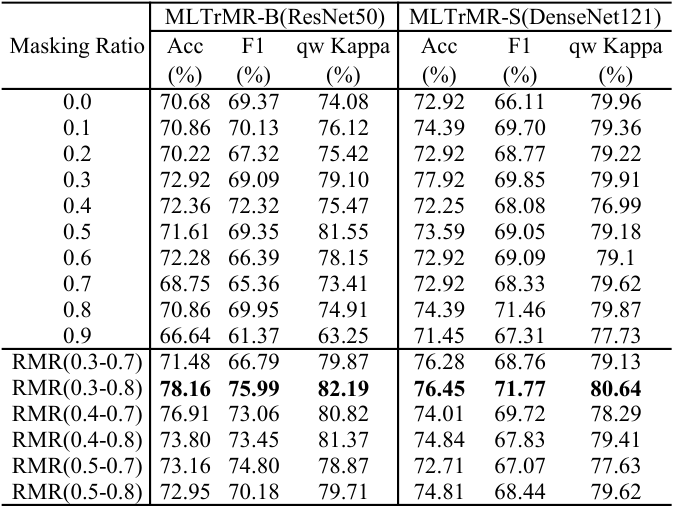}
    \label{tab:7}
\end{table}

\boldparagraph{Efficacy of the Random Masking Ratio.} We conducted experiments on the MLTrMR-B with ResNet50 as the backbone and the MLTrMR-S with DenseNet121 as the backbone to investigate the effect of various masking ratios on the model and the effectiveness of the random masking ratio. The experimental results are shown in Table~\ref{tab:7}. We observed that in the random masking ratio range of 0.3 to 0.8, i.e., RMR(0.3-0.8), both MLTrMR-B(ResNet50) and MLTrMR-S(DenseNet121) achieved optimal performance.

\boldparagraph{Efficacy of the Latent Embedder.} Table~\ref{tab:8} (a) shows that the performance of MLTrMR significantly decreases when the latent embedder is not utilized compared to incorporating the latent embedder. It can be attributed to the absence of the latent, which renders the model akin to ViT. Consequently, the inductive bias hinders the extraction of effective lesion features from the limited dataset.

\boldparagraph{Efficacy of Auxiliary Loss Function.} Table~\ref{tab:8} (b) shows that the performance of MLTrMR is better with the auxiliary loss function during training than without it. This is because masked tokens predicted by unmasked tokens in the decoder output are essentially learned features. The discrepancy between these predicted tokens and the original features serves as a measure of the quality of the model and, to some extent, influences the direction of parameter updates. Consequently, minimizing the discrepancy between these features during training can steer the model towards more favorable parameter updates, significantly improving model performance.

\boldparagraph{Efficacy of adaLN Linear.} Table~\ref{tab:8} (c) shows that adding an adaLN linear layer after the decoder can improve model performance when using auxiliary loss. The adaLN dynamically adjusts the normalization parameters according to the features of individual samples, independent of statistical information from other samples. This dynamic adjustment allows for more flexible adaptation to the feature distributions of different samples, ultimately improving model performance.

\boldparagraph{Efficacy of RelPos-MSA.} Table~\ref{tab:8} (d) shows the significant impact of relative positional bias in RelPos-MSA on improving model performance. The visualization results are shown in Fig.~\ref{fig:xu9}. Relative positional bias is essential in capturing the relationships between tokens, strengthening the model's ability to model latent features. Relative positional bias improves the model's understanding of the contextual relationships between tokens, leading to a more accurate capture of the latent features and improving model performance.

\boldparagraph{Efficacy of Maked shortcut.} Table~\ref{tab:8} (e) shows a notable difference in performance between scenarios where the masked shortcut is used and those where it is not. The masking shortcut ensures that the decoder only predicts the masked tokens, leaving the unmasked tokens. It increases model robustness and allows for better adaptation to unknown information and noise, improving performance.

\boldparagraph{Efficacy of the Masking Operation.} Table~\ref{tab:8} (f) shows that the masking operation can improve the robustness of the model. Given the limited number of images in DFID, the masking operation enhances the model's learning capacity with limited samples. It allows the model to focus on critical features, reducing excessive learning from noise or irrelevant information, thus improving its generalization performance.

\section{Conclusion}
In this paper, we construct the first open-source dataset of dental fluorosis images (DFID), laying the foundation for deep learning to diagnose dental fluorosis. In addition, we take the lead in applying deep learning to diagnosing dental fluorosis by proposing the masked latent transformer with the random masking ratio (MLTrMR) for automated intelligent diagnosis of dental fluorosis. MLTrMR introduces a mask latent modeling scheme based on Vision Transformer to enhance contextual learning of dental fluorosis lesion features. To improve the model's performance, we implemented an auxiliary loss function during the training of MLTrMR to guide parameter updates. The outstanding performance of MLTrMR on DFID makes it state-of-the-art (SOTA). In the future, our research will further explore the automated intelligent diagnosis of dental fluorosis and its application in clinical practice.

\bibliographystyle{IEEEtran}
\bibliography{references}

\end{document}